\documentclass{article}

\PassOptionsToPackage{numbers, compress}{natbib}
\usepackage[preprint]{neurips_2024}
\usepackage[utf8]{inputenc} 
\usepackage[T1]{fontenc}    
\usepackage{hyperref}       
\usepackage{url}            
\usepackage{booktabs}       
\usepackage{amsfonts}       
\usepackage{nicefrac}       
\usepackage{microtype}      
\usepackage{xcolor}         
\usepackage{graphicx}
\usepackage{appendix}
\usepackage{array}

\definecolor{darkpurple}{rgb}{0.6, 0.0, 0.6}

\title{What Are Large Language Models Mapping to in the  Brain? A Case Against Over-Reliance on Brain Scores}

\author{%
   Ebrahim Feghhi$^{* 1,2}$ \; Nima Hadidi$^{*1,2}$ \\
   \AND Bryan Song$^{1}$ \; Idan A. Blank$^{\dagger 2,3,4}$ \; Jonathan C. Kao$^{\dagger 1,2}$
}

\begin{document}
\footnote{$^*$Equal contribution, list order is random. $^\dagger$Co-senior authors. All authors are affiliated with the University of California, Los Angeles. $^1$Department of Electrical and Computer Engineering. $^2$Neuroscience Interdepartmental Program. $^3$Department of Psychology. $^4$Department of Linguistics. Contact \href{mailto:nhadidi@g.ucla.edu}{nhadidi@g.ucla.edu} or \href{mailto:ebrahimfeghhi@g.ucla.edu}{ebrahimfeghhi@g.ucla.edu} for correspondence.}
\maketitle

\begin{abstract}
Given the remarkable capabilities of large language models (LLMs), there has been a growing interest in evaluating their similarity to the human brain. One approach towards quantifying this similarity is by measuring how well a model predicts neural signals, also called "brain score". Internal representations from LLMs achieve state-of-the-art brain scores, leading to speculation that they share computational principles with human language processing. This inference is only valid if the subset of neural activity predicted by LLMs reflects core elements of language processing. Here, we question this assumption by analyzing three neural datasets used in an impactful study on LLM-to-brain mappings, with a particular focus on an fMRI dataset where participants read short passages. We first find that when using shuffled train-test splits, as done in previous studies with these datasets, a trivial feature that encodes temporal autocorrelation not only outperforms LLMs but also accounts for the majority of neural variance that LLMs explain. We therefore caution against shuffled train-test splits, and use contiguous test splits moving forward. Second, we explain the surprising result that untrained LLMs have higher-than-expected brain scores by showing they do not account for additional neural variance beyond two simple features: sentence length and sentence position. This undermines evidence used to claim that the transformer architecture biases computations to be more brain-like. Third, we find that brain scores of trained LLMs on this dataset can largely be explained by sentence length, sentence position, and static word embeddings; a small, additional amount is explained by sense embeddings and contextual representations of sentence structure. We conclude that over-reliance on brain scores can lead to over-interpretations of similarity between LLMs and brains, and emphasize the importance of deconstructing what LLMs are mapping to in neural signals.
\end{abstract}

\section{Introduction}
Recent developments in large language models (LLMs) have led many to wonder whether LLMs process language like humans do. Whereas LLMs acquire many abstract linguistic generalizations, it remains unclear to what extent their internal machinery bears resemblance to the human brain \citep{Mahowald2024-vk}. A number of studies have attempted to answer this question through the framework of neural encoding \citep{Schrimpf2021-pg,Toneva2022-lo, Caucheteux2022-dt}. Within this framework, an LLM's internal representations of some linguistic stimuli are used to predict brain activity during comprehension of the same stimuli. Results have been uniformly positive, showing that LLM representations are highly effective at predicting neural signals \citep{Jain2018-rg, Toneva2019-xy}.

In one impactful study, authors evaluated the brain scores of 43 models on three neural datasets \citep{Schrimpf2021-pg}. They found that GPT2-XL \citep{Radford2019-gr} achieved the highest brain score and, in one neural dataset, accounted for $100$\% of the "explainable" neural variance (i.e., taking into account the noise inherent in the data) \citep{Pereira2018-ry}. This result was interpreted as evidence that the brain may be optimizing for the same objective as GPT2, namely, next-word prediction. Surprisingly, the authors further found that untrained (i.e. randomly initialized) LLMs predict neural activity well, leading to speculations that the transformer architecture biases computations to be more brain-like. The finding that untrained LLMs predict neural signals significantly above chance has been replicated in other studies \citep{Pasquiou2022-rn, Caucheteux2022-dt, Hosseini2024-kg}. 

More generally, many studies have compared models to brain activity and concluded that high prediction performance reveals correspondence between some interesting aspect of the model and biological linguistic processing \citep{Caucheteux2022-dt, aw2024instructiontuned, Caucheteux2023-yf, goldstein2024the, Tikochinski2024}. One issue with this approach is that it assumes that the subset of neural activity predicted by a model reflects core processes of the human language system \citep{Bowers2023-bl}. However, this assumption is not necessarily true. For example, a recent paper found that when participants listen to stories, the fMRI signal includes an initial ramping, positional artifact \citep{Antonello2023-ab}. It is likely that LLMs which contain absolute positional embeddings would be able to predict this ramping signal, whereas a simpler model such as static word embeddings (e.g. GloVe, \citep{Pennington2014-ow}) would not, leading to exaggerated differences between LLMs and GloVe due to reasons of little theoretical interest.  This issue relates to a more general trend in machine learning research: a complex algorithm solves a task, but it is later discovered that the key innovation was a very simple component of the algorithm \citep{weinberger2024ab}. Analogous to \citet{weinberger2024ab}, we argue that without attempting to rigorously deconstruct the mapping between LLMs and brains, it is possible to draw erroneous conclusions about the brain's mechanisms for processing language.
 
We analyze the same three neural datasets used in \citet{Schrimpf2021-pg}. These include the Pereira fMRI dataset, where participants read short passages \citep{Pereira2018-ry}; the Fedorenko electrocorticography (ECoG) dataset, where participants read isolated sentences \citep{Fedorenko2011-kd}; and the Blank fMRI dataset, where participants listened to short stories \citep{Blank2014-iz}. As in \citet{Schrimpf2021-pg}, we focus our analyses on \textit{Pereira}. In order to deconstruct the mapping between LLMs and the brain, we follow \citet{Reddy2021-rd} and \citet{De_Heer2017-rl} by building a set of simple features of the linguistic input, and gradually adding features of increasing complexity. Our goal is to find the simplest set of features which account for the greatest portion of the mapping between LLMs and brains. 

\section{Methods}

\subsection{Experimental data}
For all three datasets, we used the same versions as used by \citep{Schrimpf2021-pg}. For additional details, refer to \ref{sec:experimental_data}.

\textbf{\textit{Pereira} (fMRI):} \textit{Pereira} is composed of two experiments. Experiment 1 (EXP1) consists of $96$ passages each containing $4$ sentences, with $n=9$ participants. Experiment 2 (EXP2) consists of $72$ passages each consisting of $3$ or $4$ sentences, with $n=6$ participants. Passages in each experiment were evenly divided into $24$ semantic categories which were not related across experiments ($4$ passages per category in EXP1, and $3$ passages per category in EXP2). A single fMRI scan (TR) was taken after visual presentation of each sentence. Unless otherwise noted, we focus our results on voxels from within the "language network" in the main paper \citep{Fedorenko2024-jt}. EXP1 was a $384 \times 92450$ matrix (number of sentences $\times$ number of voxels) and EXP2 was a $243$ $\times$ $60100$ matrix. All analyses were conducted separately for each experiment, as in previous studies \citep{Schrimpf2021-pg, aw2023training, Kauf2024-rh, Hosseini2024-kg}. 

\textbf{\textit{Fedorenko} (ECoG): } Participants ($n=5$) read $52$ sentences of length $8$ words ($416$ words).  A total of $97$ language-responsive electrodes were used across $5$ participants: $47, 8, 9, 15,$ and $18$, for participants $1$ through $5$, respectively. High gamma activity was extracted from the neural recordings, and responses were temporally averaged  across the full presentation of each word. The entire dataset was a $416 \times 97$ matrix.

\textbf{\textit{Blank} (fMRI): } \textit{Blank} consisted of $5$ participants listening to $8$ stories from the publicly available Natural Stories Corpus \citep{Futrell2018-sk}. An fMRI scan was taken every $2$ seconds, resulting in a total of $1317$ TRs across the $8$ stories. fMRI BOLD signals were averaged across voxels within each functional region of interest (fROI). There were $60$ fROIs across all $5$ participants, resulting in a $1317 \times 60$ matrix.

\raggedbottom

\subsection{Language models} 
We focus our analyses on \textit{GPT2-XL} \citep{Radford2019-gr}, as it was shown to be the best-performing model on \textit{Pereira}  \citep{Hosseini2024-kg, Kauf2024-rh, Schrimpf2021-pg}. \textit{GPT2} is an auto-regressive transformer model, meaning that it can only attend to current and past tokens, trained on next token prediction. The \textit{XL} variant has $\sim1.5$B parameters and $48$ layers. We replicate our findings with trained \textit{GPT2-XL} on \textit{Pereira} with \textit{RoBERTa-Large}\citep{Liu2019-ej} (\ref{sec:roberta-large}). \textit{RoBERTa} is a transformer model with bidirectional attention, meaning that it can attend to past and future tokens, trained on masked token prediction. The large variant contains $\sim335$M parameters and $24$ layers. Both \textit{GPT2} and \textit{RoBERTa} use learned absolute positional embeddings, such that a unique vector corresponding to each token position is added to the input static embeddings. 
\subsection{Description of feature spaces}
We provide a description of the feature spaces used to account for the neural variance explained by LLMs in Table \ref{tab:FEATURES}. 
\begin{table}[h]
    \centering
    \caption{Description of feature spaces: sentence position (SP), sentence length (SL), static word embeddings (WORD), sense embeddings (SENSE), syntactic embeddings (SYNT), and word position (WP). The Dataset column indicates which dataset the feature space was used for. Additional details are provided in linked sections to the appendix.}
    \begin{tabular}{ m{1cm} m{8.3cm} m{0.8cm} m{1cm}} 
      \hline
       \toprule
      \textbf{Feature space} & \multicolumn{1}{c}{\textbf{Description}} & \textbf{Size} & \textbf{Dataset} \\ \bottomrule
      \hline 
      SP & One-hot vector where each element corresponds to a given position within a passage. & $4D$ & Pereira \\ \hline
      SL & Number of words in each sentence. & $1D$ & Pereira \\  \hline
      WORD & Static representation of each content word (i.e. nouns, verbs, adjectives, adverbs) generated on pronoun dereferenced text and sum-pooled within each sentence (\ref{sec:Static}). & $1024D$ & Pereira \\ \hline
      SENSE & Sense-specific representation of each content word generated on pronoun dereferenced text and sum-pooled within each sentence (\ref{sec:LMMS}). & $1024D$ & Pereira \\ \hline
      SYNT & Syntactic representation of each sentence, generated by averaging LLM representations (from the best layer) across $100$ sentences with equivalent syntactic structure but different semantic content (\ref{sec:SYNT}). & $1600D$ & Pereira \\ \hline
      WP & A ramping positional signal combined with a positional signal which encodes nearby words similarly (\ref{sec:FedWP}). & $9D$ & Fedorenko
    \end{tabular}
    \label{tab:FEATURES}
\end{table}

\subsection{LLM feature pooling} 
\label{llm feature pooling}
\textbf{\textit{Pereira}:} Each sentence was fed into an LLM, with previous sentences from the same passage also fed as input. Since each fMRI scan was taken at the end of the sentence, we converted LLM token-level embeddings to sentence-level embeddings by summing across all tokens within a sentence (sum-pooling). We used sum-pooling because it performed similarly or better than taking the representation at the last token (\ref{sec:across_layer}), as done in previous analyses of this dataset\citep{Schrimpf2021-pg, Hosseini2024-kg, Kauf2024-rh, aw2024instructiontuned}, and because it is more analogous to the word-level impulse response model commonly used in more naturalistic neural encoding studies. \citep{Huth2016-tp,Jain2020-lo},

\textbf{\textit{Fedorenko}: } The current and previous tokens from within the same sentence were fed into the LLM as context. We converted LLM token-level embeddings to word embeddings, since each word has a recorded neural response, by summing across tokens in multi-token words, and leaving single token words unmodified.

\textbf{\textit{Blank}:} For each story, we fed the current and all preceding tokens up to a maximum context size of $512$ tokens. As in \citet{Schrimpf2021-pg}, for each TR, we took the representation of the word that was closest to being $4$ seconds before the TR. For multi-token words, we took the representation of the last token of that word. 

\subsection{Banded ridge regression}
We use ridge regression (linear regression with an L2 penalty) to predict activations for each voxel/electrode/fROI independently. When fitting regressions with both large and small feature spaces, we employ banded ridge regression to give small feature spaces their own L2 penalty \citep{Dupre_la_Tour2022-sy} \ref{sec:banded ridge}. This is because "vanilla" ridge regression can be biased against making use of small feature spaces. 

\subsection{Out of sample $R^2$ metric}
We define the brain score of a model (set of feature spaces used in a regression) as the out-of-sample $R^2$ metric ($R^2_{oos}$) \citep{Hawinkel_undated-jn}. $R^2_{oos}$ quantifies how much better a set of features performs at predicting held-out data compared to a model which simply predicts the mean of the training data (i.e. a regression with only an intercept term). To be precise, given mean squared error (MSE) values from a model \textit{M} and MSE values from an intercept-only regression (\textit{I}), then:
\begin{eqnarray}
R^2_{oos} = 1 - \frac{MSE_{M}}{MSE_{I}}.
\end{eqnarray}
A positive (negative) value indicates that $M$ was more (less) helpful than predicting the mean of training data. We elected to use $R^2_{oos}$ over the standard $R^2$ because of this clear interpretation and because it is a less biased estimate of test set performance \citep{Hawinkel_undated-jn}. We use $R^2_{oos}$ over Pearson's correlation coefficient ($r$) because $R^2_{oos}$ can be interpreted as the fraction of variance explained, which lends more straightforwardly to estimating how much variance one feature space explains over others. Whenever averaging across voxels, we set $R^2_{oos}$ values to be non-negative to prevent differences in performance on noisy voxels/electrodes/fROIs from significantly impacting the results. We refer to $R^2_{oos}$ as $R^2$ throughout the rest of the paper for brevity, and use the notation $R^2$\textsubscript{\textit{M}} to refer to the performance of model \textit{M}.  

\subsection{Selection of best layer}
We evaluate the $R^2$ for each LLM layer, and select the layer that performs best across voxels/electrodes/fROIs. Due to the stochastic nature of untrained LLMs, we selected the best layer for $10$ random seeds and report the average $R^2$ across seeds. When reporting the best layer, we refer to layer $0$ as the input static layer, and layer $1$ as the first intermediate layer. 

\subsection{Train, validation, and test folds:}
For each dataset, we construct contiguous train-test splits by ensuring neural data from the same passage/sentence/story is not included in both train and test data. Due to low sample sizes, we employed a nested cross-validation procedure for each dataset (\ref{sec:train_test}). When computing $R^2$ across inner or outer folds, we pooled predictions across folds and computed a single $R^2$  as recommended by \citet{Hawinkel_undated-jn}. The optimal hyperparameters were selected based on validation data. We created shuffled train-test splits, as done in \cite{Schrimpf2021-pg}, of the same size as the contiguous train-test splits. Unless explicitly noted, all results are performed using contiguous train-test splits. 

\subsection{Correcting for decreases in test-set performance due to addition of feature spaces}
\label{sec:quantifying_overlap}
It is possible for a full encoding model to perform worse than a sub-model (which consists of only a subset of the feature spaces) due to overfitting \citep{De_Heer2017-rl}. For instance, it is possible for a joint regression with an LLM and additional feature spaces to perform worse than a model containing the LLM alone. To address this problem, in some analyses we select the best performing sub-model for each voxel/electrode/fROI, and denote this model with a $*$. More precisely, the corrected value for $R^2$\textsubscript{\textit{A+B+C}} is: 

\begin{eqnarray}
    R^2\textsubscript{\textit{A+B+C}}* = \max({R^2\textsubscript{\textit{A}}, R^2\textsubscript{\textit{B}}, R^2\textsubscript{\textit{C}}, R^2\textsubscript{\textit{A+B}}, R^2\textsubscript{\textit{A+C}},   R^2\textsubscript{\textit{B+C}}, R^2\textsubscript{\textit{A+B+C}}}).
\label{eqn:max}
\end{eqnarray}
Since the goal of this study is to deconstruct the mapping between LLMs and brains, we force the max procedure in Eq. \ref{eqn:max} to include the LLM when using a model with the best layer of an LLM as a feature space. We denote this model as $R^2$\textsubscript{\textit{M+\textbf{LLM}}}*, where the bold indicates the LLM is not removed. 

\subsection{Percentage of LLM neural predictivity accounted for}
\label{sec:frac_var_llm}
Given a model $M$, we quantify the percentage of neural variance explained by an LLM that is also explained by $M$ through Eq. \ref{eqn:vaf}.

\begin{eqnarray}
    \Omega \textsubscript{\textit{LLM}}(\textit{M}) = \left(1 - \frac{R^2\textsubscript{\textit{M}+\textbf{\textit{LLM}}}* - R^2\textsubscript{\textit{M}}*}{R^2\textsubscript{\textit{LLM}}}\right) \times 100\%
    \label{eqn:vaf}
\end{eqnarray}
For a given voxel/electrode/fROI, 
 $\Omega \textsubscript{\textit{LLM}}(\textit{M})$ is less than (more than) $100\%$ if the best sub-model that includes the LLM outperforms (underperforms) the best sub-model that does not include the LLM. When reporting this quantity, we average across voxels/electrodes/fROIs within each participant, clip the within-participant averages to be at most equal to $100\%$ to prevent noisy estimates from leading to an upward bias, and then report the mean $\pm$ standard error of the mean (SEM) across participants. 

\subsection{Orthogonal Autocorrelated Sequences Model (\textit{OASM})}
To model temporal autocorrelation in neural activity, we construct a feature space for each dataset by (i) forming an $n$-dimensional identity matrix, where $n$ is the total number of time points in the dataset, and (ii) applying a Gaussian filter within blocks along the diagonal that correspond to temporally contiguous time points (i.e., within each passage in \textit{Pereira}, each sentence in \textit{Fedorenko}, and each story in \textit{Blank}). This generates an autocorrelated sequence for each passage/sentence/story that is orthogonal to that of each other passage/sentence/story (\ref{sec:OASM}). 

We computed the unique neural variance explained by an LLM relative to what \textit{OASM} explains alone through Eq. \ref{eqn:unqvar}. When $\Phi$\textsubscript{\textit{LLM}} is below $100\%$, any potentially linguistically-driven neural variance explained uniquely by the LLM is less than the neural variance explained by a model with no linguistic features. Values are reported in the same manner as $\Omega$, except that no clipping is performed.

\begin{eqnarray}
\label{eqn:unqvar}
 \Phi\textsubscript{\textit{LLM}} = \left(\frac{R^2\textsubscript{\textit{OASM}+\textit{\textbf{LLM}}}*}{R^2\textsubscript{\textit{OASM}}} - 1\right) \times 100\%
\end{eqnarray}

\section{\textit{Pereira}}

\subsection{Shuffled train-test splits are severely affected by temporal autocorrelation}
\begin{figure}[t]
    \centering
    \includegraphics[scale=0.89]{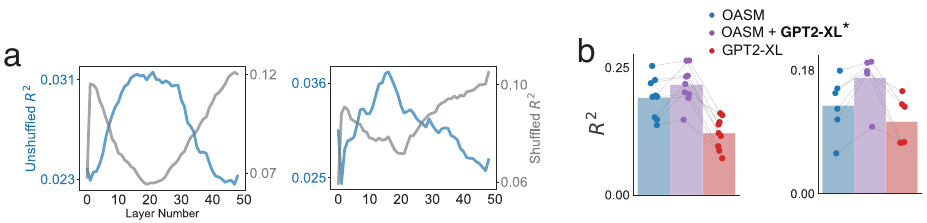}
    \caption{Within each panel, EXP1 results are on the left and EXP2 results are on the right \textbf{(a)} $R^2$ values across layers for \textit{GPT2-XL} on shuffled train-test splits (gray) and contiguous (unshuffled) splits (blue). \textbf{(b)} Each dot shows the mean $R^2$ value across voxels within a participant, with bars indicating mean $R^2$ across participants.}
    \label{fig:shuffled_pereira}
\end{figure}

Prior LLM encoding studies using this dataset \citep{Kauf2024-rh, Schrimpf2021-pg, Hosseini2024-kg, Oota2022-bc, aw2024instructiontuned} used shuffled train-test splits. Here, we demonstrate that this approach compromises the evaluation of the neural predictivity of LLMs. First, we replicated the pattern of neural predictivity across \textit{GPT2-XL}'s layers reported in \citep{Schrimpf2021-pg} \footnote{Though the pre-computed scores provided by \citet{Schrimpf2021-pg} at \href{https://github.com/mschrimpf/neural-nlp/tree/master}{neural-nlp} exhibit the pattern described here, Fig. $2$ in their main paper does not for reasons we are unaware of.} and \citep{Kauf2024-rh} when using shuffled splits; early and late layers perform best and intermediate layers perform worst. Strikingly, when using contiguous train-test splits, the opposite pattern is observed: intermediate layers perform best. Across layers, the neural predictivity found when using shuffled splits is highly anti-correlated with that found when using contiguous splits  ($r=-.929$ in EXP1, $r=-.764$ in EXP2) (Fig. \ref{fig:shuffled_pereira}a), showing that shuffled splits can yield contradictory results to contiguous splits.

Next, we hypothesized that much of what LLMs map to when using shuffled splits could be accounted for by \textit{OASM}, a model which only represents within passage autocorrelation and between passage orthogonality. \textit{OASM} outperformed \textit{GPT2-XL} on both EXP1 and EXP2 (Fig. \ref{fig:shuffled_pereira}b, blue and red bars), revealing that a non-linguistic feature space can achieve absurdly high brain scores in the context of shuffled splits. We next examined the fraction of $R^2$\textsubscript{\textit{GPT2-XL}} that is also explainable by \textit{OASM} by fitting a regression with both \textit{OASM} and \textit{GPT2-XL}. After correcting for voxels where adding \textit{OASM} weakens the performance of \textit{GPT2-XL} (denoted \textit{OASM} $+$ \textit{\textbf{GPT2-XL}}*, \ref{sec:quantifying_overlap}), we find that the majority of $R^2$\textsubscript{\textit{GPT2-XL}} is explained by \textit{OASM}: $\Omega \textsubscript{\textit{GPT2-XL}}(\textit{OASM})=$  $81.5\pm 5.5\%$ in EXP1, $62.7\pm4.7\%$ in EXP2 (mean $\pm$ SEM across participants). Moreover, we find that the unique neural variance \textit{GPT2-XL} explains relative to what \textit{OASM} explains alone is very small: $\Phi$\textsubscript{\textit{GPT2-XL}} $= 15.3\pm5.5\%$ in EXP1 and $37.7\pm9.0\%$ in EXP2. Thus, when using shuffled train-test splits, it appears that the largest determinant of model predictivity on this dataset is whether a model contains autocorrelated sequences within passages that are orthogonal between passages. This strongly challenges the assumption of multiple previous studies \citep{Schrimpf2021-pg, aw2024instructiontuned, Hosseini2024-kg} that performance on this benchmark is an indication of a model's brain-likeness (see \ref{sec:shuffled} for further discussion). 
\raggedbottom

\subsection{Untrained \textit{GPT2-XL} neural predictivity is fully accounted for by sentence length and position}

We next sought to deconstruct what explains the neural predictivity of untrained \textit{GPT2-XL} (\textit{GPT2-XLU}) in \textit{Pereira}. We hypothesized that $R^2$\textsubscript{\textit{GPT2-XLU}} could be explained by two simple features: sentence length (\textit{SL}) and sentence position within the passage (\textit{SP}). Sentence length is captured by \textit{GPT2-XLU} because the GELU nonlinearity in the first layer's MLP transforms normally distributed inputs with zero mean into outputs with a positive mean. This introduces a positive mean component to each token’s representation in the residual stream. When these representations are sum-pooled, this positive mean component accumulates in a way that reflects the sentence length, making the length decodable in the intermediate layers. Sentence position is encoded within \textit{GPT2-XLU} due to absolute positional embeddings which, although untrained, still result in sentences at the same position having similar representations when tokens are sum-pooled.

To obtain representations from \textit{GPT2-XLU}, we selected the best-performing layer for each of the $10$ untrained seeds. For EXP1 the best performing layer was layer $0$ for $6$ seeds, layer 1 for $3$ seeds (first intermediate layer), and layer $19$ for one seed. For EXP2 the best layer was layer $1$ for 5 seeds, layer $2$ for 4 seeds, and layer $5$ for $1$ seed. 

\begin{figure}[h]
    \centering
    \includegraphics[scale=0.95]{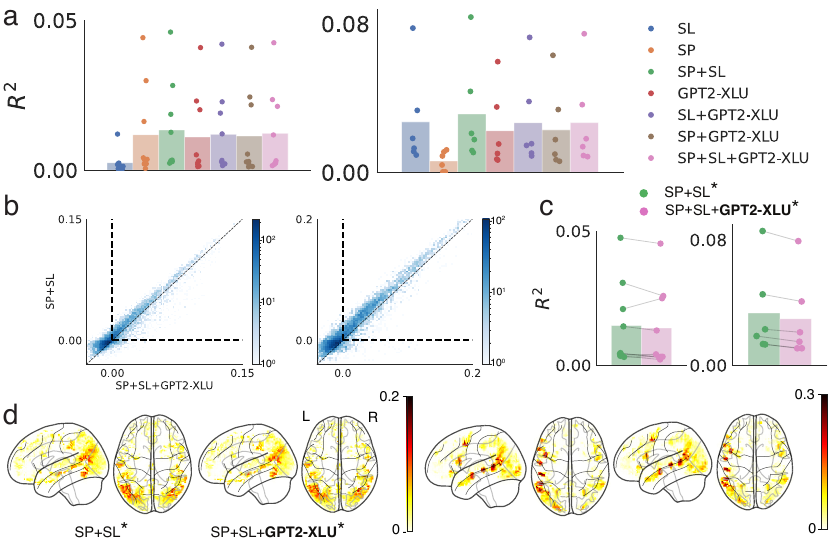}
    \caption{For all panels, EXP1 results are on the left and EXP2 results are on the right. \textbf{(a)} $R^2$ for different combinations of features. Each dot represents $R^2$ values averaged across voxels in a single participant, with bars showing mean across participants. \textbf{(b)} 2D histogram of $R^2$ values ($100$ bins) for the best model without \textit{GPT2-XLU} (\textit{SP+SL}), and the best model with \textit{GPT2-XLU} (\textit{GPT2-XLU+SP+SL}). The dotted lines show $y=x$, $y=0$, and $x=0$. Values below $y=0$ or left of $x=0$ were clipped when averaging, but are shown here to visualize the full distribution. \textbf{(c)} Same as \textbf{(a)}, but after voxel-wise correction for \textit{SP+SL} and \textit{SP+SL+GPT2-XLU}; lines connect data-points from the same participant. \textbf{(d)} Glass brain plots, for a representative participant (different for each EXP), showing $R^2$ values of \textit{SP+SL*} (left) and \textit{SP+SL+\textbf{GPT2-XLU}*} (right). Neural encoding performance across brain areas is highly similar for both models.}
    \label{fig:untrained_pereira}
\end{figure}

We fit a regression using all subsets of the following feature spaces: \{\textit{SP, SL, GPT2-XLU}\}, resulting in $7$ models. For both experiments, $R^2$\textsubscript{\textit{SP+SL}} was higher than the $R^2$ of any other model, including the best-performing model with \textit{GPT2-XLU} (\textit{SP+SL+GPT2-XLU}) (Fig. \ref{fig:untrained_pereira}a,b). Sentence position was particularly important in EXP1, and sentence length was particularly important in EXP2, which may explain why the static layer performed well in EXP1 despite not encoding sentence length as well as the intermediate layers.
Although \textit{GPT2-XLU} does not enhance encoding performance when averaging across voxels, there may be a subset of voxels where it does explain significant additional neural variance. To address this, we performed a one-sided paired $t$-test between the squared error values obtained over sentences (EXP1: $N=384$ , EXP2: $N=243$) using \textit{SP+SL+GPT-XLU} and \textit{SP+SL} (\ref{sec:stats_test}). Only $1.83$\% (EXP1) and $1.33$\% (EXP2) of voxels were significantly  ($\alpha=0.05$) better explained by the model with \textit{GPT2-XLU} before false discovery rate (FDR) correction; no voxel was significant in either experiment after performing FDR correction within each participant \citep{Benjamini1995-kn}. Taken together, these results suggest \textit{GPT2-XLU} does not enhance neural prediction performance over sentence length and position.

To control for voxels where the neural encoding performance of \textit{GPT2-XLU} is weakened by the addition of \textit{SP+SL}, we compared \textit{SP+SL*} and \textit{SP+SL+\textit{\textbf{GPT2-XLU}}*}. \textit{SP+SL} accounted for almost all neural variance explained by \textit{GPT2-XLU}: $\Omega\textsubscript{GPT2-XLU}(\textit{SP+SL}) = 98.4\% \pm 1.5\%$ in EXP1 and $100.0\% \pm 0.0\%$ in EXP2 (Fig. \ref{fig:untrained_pereira}c,d). Only $1.26$\% (EXP1) and $0.95$\% (EXP2) of voxels were significantly better explained by the addition of \textit{GPT2-XLU} before FDR correction; no voxel was explained significantly better by the addition of \textit{GPT2-XLU} after FDR correction. Thus, our results hold even when controlling for decreases in performance due to the addition of feature spaces.

\subsection{Sentence length, sentence position, and static word embeddings account for the majority of trained \textit{GPT2-XL} neural predictivity}
We next turned to explaining the neural predictivity of the trained \textit{GPT2-XL}. In addition to sentence position and sentence length, we added static word embeddings (\textit{WORD}) generated on a version of the text where pronouns are dereferenced. Together, these features defined a baseline model which does not account for any form of linguistic processing of words in context beyond pronoun resolution. We next included two more features which involved contextual processing. These were sense-specific word embeddings, which contain distinct representations for different senses of the same word (e.g., mouse: \textit{computer device}, and mouse: \textit{rodent}) \citep{Loureiro2022-ww} (\textit{SENSE}), and syntactic representations generated using a method similar to \citet{Caucheteux2021-nv} (\textit{SYNT}). We selected the best layer based on averaged $R^2$ across language voxels on test data (EXP1: layer $21$, EXP2: layer $16$). 

\begin{table}[h]
    \centering
    \caption{Mean $R^2$ values (across participants) for each model. For models composed of multiple feature spaces, the best sub-model is used which includes the last feature space. Blue (red) feature spaces are only used in EXP1 (EXP2), purple feature spaces are used in both experiments, and black feature spaces are used in neither.}
    \begin{tabular}{cccc}
         \hline
         \toprule
         \rule{0pt}{2ex} \textbf{Model}&\textcolor{blue}{\textbf{EXP1}} &\textcolor{red}{\textbf{EXP2}}&\textbf{Average}\\ \bottomrule \hline 
          \rule{0pt}{2ex} \textcolor{darkpurple}{GPT2-XL}& 
    $0.0315$&$0.0371$&$0.0343$\\ \hline
 \textcolor{darkpurple}{SP+SL}&$0.0134$&$0.0314$&$0.0224$\\ \hline
 \textcolor{darkpurple}{SP+SL+WORD}&$0.0248$&$0.0395$&$0.0322$\\ \hline
 \textcolor{darkpurple}{SP+SL+WORD+SENSE}&$0.0260$&$0.0398$&$0.0329$\\  \hline
 SP+\textcolor{darkpurple}{SL}+\textcolor{darkpurple}{WORD}+\textcolor{blue}{SENSE}+\textcolor{darkpurple}{SYNT}&$0.0268$&$0.0425$&$0.0347$\\  \hline
 SP+\textcolor{darkpurple}{SL}+\textcolor{red}{WORD}+SENSE+\textcolor{red}{SYNT}+\textcolor{darkpurple}{GPT2-XL}&$0.0319$&$0.0449$&$0.0384$\\\end{tabular}
    \label{tab:Trained results}
\end{table}

\begin{figure}[h]
    \centering
    \includegraphics[scale=0.95]{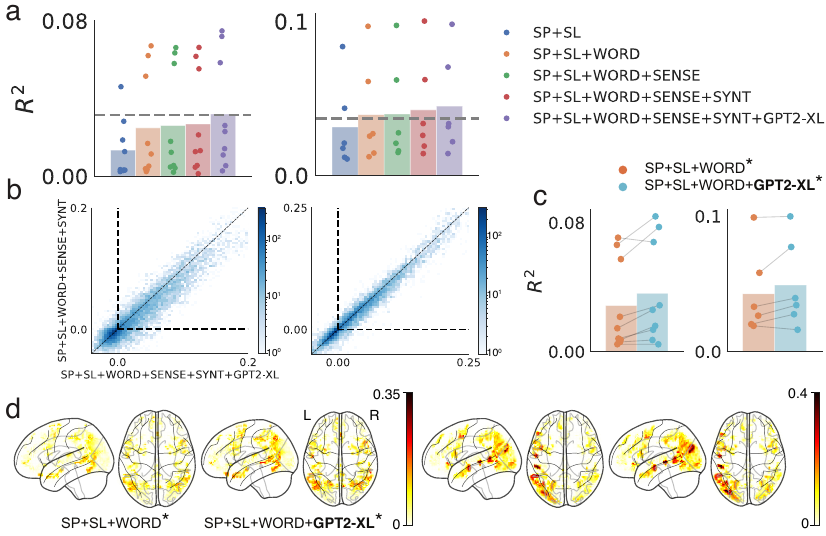}
    \caption{For all panels, EXP1 results are on the left and EXP2 results are on the right. \textbf{(a)} For each model, we display the best sub-model which must include the last feature space. Dots represent participants and bars are mean across participants. Gray dashed line is the performance of \textit{GPT2-XL} alone. \textbf{(b)} $2$D histogram comparing the best sub-model which includes \textit{GPT2-XL} to the best sub-model that does not. \textbf{(c)} Same as \textbf{(a)} but after voxel-wise correction for \textit{SP+SL+WORD} and \textit{SP+SL+WORD+GPT2-XL}. \textbf{(d)} Glass brain plots showing $R^2$ values of \textit{SP+SL+WORD*} (left) and \textit{SP+SL+WORD+\textbf{GPT2-XL}*} (right) for a representative participant (different for each EXP).}
    \label{fig:trained_pereira}
\end{figure}

We fit a regression to the fMRI data using all models with the following feature spaces: \{\textit{SL+SP, WORD, SENSE, SYNT, GPT2-XL}\}, resulting in $63$ models. In this list, features are ranked from least to most complex. For each feature space, we took the model that exhibited the best performance in the language network which included that feature space but did not include feature spaces more complex than it. For instance, $R^2$\textsubscript{\textit{SP+SL+WORD+SENSE}} is obtained by taking the best performing model which included \textit{SENSE}, excluding models which included \textit{SYNT} and \textit{GPT2-XL}. By doing so, we were able to examine the impact of adding more complex features after accounting for simpler features. Since this procedure is not performed at the voxel-level, we do not add a * to $R^2$. 

\autoref{tab:Trained results} displays the performance of each model, including \textit{GPT2-XL} on its own (Fig. \ref{fig:trained_pereira}a, \ref{fig:trained_pereira}b). The baseline \textit{SP+SL+WORD }model, which does not account for any form of contextual processing, performs $72.3\% \pm 9.2$\%  as well as \textit{GPT2-XL} in EXP1 (mean $\pm$ SEM across participants), and outperforms \textit{GPT2-XL }in EXP2. When adding \textit{SENSE} and \textit{SYNT}, the joint model performs $83.6\% \pm 9.9\%$ as well as \textit{GPT2-XL} and $78.5\% \pm 8.9\%$ as well as the full model (all feature spaces including \textit{GPT2-XL}) in EXP1, and better than \textit{GPT2-XL} and $93.3\% \pm 3.1\%$ as well as the full model in EXP2. Together, these results indicate that the baseline model performs well relative to \textit{GPT2-XL}, and that more complex forms of contextual processing, namely word sense disambiguation and contextual syntactic representation, play a modest role after taking into account simpler features.

Similar to previous sections, we perform voxel-wise correction by selecting the best sub-model with \textit{GPT2-XL} and the best sub-model without \textit{GPT2-XL} for each voxel. We once again operate in a layered fashion by starting with simple features and working up in complexity. Table \ref{tab:Trained results vc} shows the results for each joint model with and without \textit{GPT2-XL}. \textit{SP+SL+WORD*} accounts for the majority of the neural variance predicted by \textit{GPT2-XL } (Fig. \ref{fig:trained_pereira}c), and the spatial distribution of neural variance explained by SP+SL+WORD*across brain areas was similar to that of \textit{SP+SL+WORD+\textbf{GPT2-XL}}* (Fig. \ref{fig:trained_pereira}d). \textit{SENSE} and \textit{SYNT} account for an additional, modest portion of neural variance explained by \textit{GPT2-XL} (less than $10\%$ in both experiments). Hence, a model with only one form of contextual processing, namely pronoun resolution, accounts for the majority of neural variance explained by \textit{GPT2-XL} on this dataset. 

We generated static word embeddings by averaging across contextual embeddings from a transformer to maintain consistency with sense embeddings (\ref{sec:Static}). To ensure our findings were not contingent on transformer based static embeddings, we replaced \textit{WORD} in the baseline model with static embeddings from \textit{GloVe} (\ref{sec:GloVe}). Results were consistent when using \textit{GloVe}: $\Omega$\textsubscript{\textit{SP+SL+GloVe}}(\textit{GPT2-XL}) $=82.3\% \pm 6.9\%$ in EXP1, and $=87.0\% \pm 5.2\%$ in EXP2 (Fig. \ref{fig:glove}a). We further sought to simplify our baseline model by evaluating whether whether pronoun resolution was a critical computation to explain \textit{GPT2-XL} neural encoding performance. In order to do so, we generated \textit{GloVe} embeddings on text that was not pronoun dereferenced (\textit{GloVe-NPD}). $\Omega$\textsubscript{\textit{SP+SL+GloVe-NPD}}(\textit{GPT2-XL}) $=85.4\% \pm 4.7\%$ in EXP1, and $=90.1\% \pm 3.7\%$ in EXP2, indicating that a model with no contextual processing accounts for the majority of neural variance explained by \textit{GPT2-XL} in \textit{Pereira} (Fig. \ref{fig:glove}b).

\begin{table}[t]
    \centering
    \caption{From left to right: model name, mean $R^2$ across participants after voxel correction, after voxel correction with \textit{GPT2-XL} (ensuring \textit{GPT2-XL} is included as a feature space), and the percentage of \textit{GPT2-XL} neural variance explained by the model ($\Omega\textsubscript{\textit{GPT2-XL}}(\textit{M}))$. Results are shown as EXP1, EXP2.}
    \begin{tabular}{cccc}
         \toprule
         \rule{0pt}{2ex} \textbf{Model} & \textbf{Voxel-corrected} & \textbf{w/ GPT2-XL} & \textbf{$\Omega\textsubscript{\textit{GPT2-XL}}(\textit{M})$} \\ 
         \bottomrule 
         \hline 
         \rule{0pt}{2ex} SP+SL & $0.0147$, $0.0333$ & $0.0340$, $0.0477$ & $39.1 \pm 7.8$, $60.7 \pm 10.2$\\ 
         \hline
         +WORD & $0.0285$, $0.0478$ & $0.0362$, $0.0528$ & $81.2 \pm 5.2$, $90.1 \pm 4.1$\\ 
         \hline
         +SENSE & $0.0324$, $0.0508$ & $0.0377$, $0.0548$ & $86.2 \pm 3.8$, $92.5 \pm 2.8$\\ 
         \hline
         +SYNT & $0.0354$, $0.0576$ & $0.0388$, $0.0579$ & $89.9 \pm 3.8$, $97.8 \pm 1.5$\\   
    \end{tabular}
    \label{tab:Trained results vc}
\end{table}

\section{\textit{Fedorenko}}
\subsection{Shuffled train-test splits also impact \textit{Fedorenko}, but less than with \textit{Pereira}}
\begin{figure}[h]
    \centering
    \includegraphics[scale=0.8]{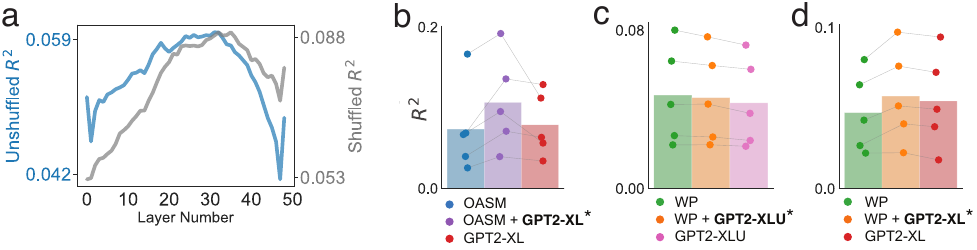}
    \caption{\textbf{(a)} Across-layer $R^{2}$, averaged across electrodes in \textit{Fedorenko}, for \textit{GPT2-XL} with and without shuffled splits. \textbf{(b, c, d)} Dots represent participants and bars are mean across participants.
    }
    \label{fig:fed}
\end{figure}

We first evaluated the impact of shuffled train-test splits on \textit{Fedorenko}. Unlike in \textit{Pereira}, the across-layer performance is well correlated between shuffled and contiguous splits ($r=0.622$) (Fig. \ref{fig:fed}a). Nonetheless, \textit{OASM} performs $91.6\pm 10.9$\% as well as \textit{GPT2-XL} (Fig. \ref{fig:fed}b), and $\Omega \textsubscript{\textit{GPT2-XL}}(\textit{OASM})=56.8\pm4.9\%$, indicating that within-sentence autocorrelation accounts for about half of the variance explained by \textit{GPT2-XL} when using shuffled splits. Furthermore, $\Phi$\textsubscript{\textit{GPT2-XL}} was $57.8 \pm 12.8$\%, meaning that any potentially linguistically-driven neural variance explained by \textit{GPT2-XL} is roughly half of what \textit{OASM} explains alone. Therefore, shuffled train-test splits also impact results on \textit{Fedorenko}, albeit less than in \textit{Pereira}, which may be due to lower autocorrelation of ECoG compared to fMRI. We use contiguous splits for the remainder of \textit{Fedorenko} analyses. 
\subsection{Word position explains all of untrained, and most of trained, \textit{GPT2-XL} neural predictivity}
As noted in \citep{Fedorenko2016-eg}, there was a strong positional signal in the ECoG dataset during comprehension of sentences that is likely related to the construction of sentence meaning. We therefore hypothesized that a feature space that accounted for word position (\textit{WP}) would do well relative to untrained and trained \textit{GPT2-XL}. We generated a simple feature space that encodes word position, such that words in nearby positions were given similar representations (\ref{sec:FedWP}). When performing a one-sided paired $t$-test between the squared errors of \textit{WP} + \textbf{\textit{GPT2-XLU}}* and \textit{WP}, $3$ electrodes were significantly better explained by the addition of \textit{GPT2-XLU} before FDR correction, and none were better explained after within-participant FDR correction (Fig. \ref{fig:fed}c). For trained \textit{GPT2-XL}, \textit{WP} performs $90.8 \pm 8.0$\% as well as \textit{GPT2-XL}, and $\Omega\textsubscript{\textit{GPT2-XL}}(\textit{WP}) =$ $81.6\pm4.5\%$ (Fig. \ref{fig:fed}d). Our results therefore suggest that the mapping between \textit{GPT2-XL} and neural activity on \textit{Fedorenko} is largely driven by positional signals. 

\section{\textit{Blank}}
\subsection{\textit{Blank} is predicted at chance levels}
Lastly, we address \textit{Blank}. We find that \textit{OASM} achieves an $R^2$ that is $103.6$ times larger than that of \textit{GPT2-XL} when using shuffled splits (\ref{sec:BlankOASM}), demonstrating that such splits are massively contaminated by temporal autocorrelation. We next turn to using contiguous splits, and test whether \textit{GPT2-XL} performs better than an intercept only model by applying a one-sided paired $t$-test between the squared error values obtained from \textit{GPT2-XL} and the intercept only model ($N=1317$ TRs). \textit{GPT2-XL} predicts $1$ fROI significantly better than an intercept only model before FDR correction, and $0$ fROIs are significantly predicted after applying FDR correction within participants. Our results therefore show that \textit{GPT2-XL} performs at chance levels on the version of \textit{Blank} used by \citep{Schrimpf2021-pg, Hosseini2024-kg, aw2024instructiontuned}. 

\section{Limitations and Conclusions}
Our study has three main limitations. First, our method of examining how much neural variance an LLM predicts over simple features scales poorly when the number of feature spaces is large. This is because additional feature spaces increases the search space of optimal regularization penalties in banded regression exponentially. Second, although we applied voxel-wise correction and banded regression to mitigate cases where adding feature spaces decreases performance, it is still possible our procedure is biased against regressions with many feature spaces due to noise in neural data and low sample sizes. Finally, we did not analyze datasets with large amounts of neural data per participant, for instance \citep{LeBel2023-ck}, in which the gap between the neural predictivity of simple and complex features may be much larger.

In summary, we find that on \textit{Pereira}, shuffled splits are heavily impacted by temporal autocorrelation, untrained \textit{GPT2-XL} brain score is explained by sentence length and position, and trained \textit{GPT2-XL} brain score is largely explained by non-contextual features. On \textit{Fedorenko}, we find that all of \textit{GPT2-XLU} brain score and the majority of \textit{GPT2-XL} brain score is accounted for by word position, and on \textit{Blank} \textit{GPT2-XL} predicts neural activity at chance levels. These results suggest that (i) brain scores on these datasets should be interpreted with caution, and (ii) more generally, analyses using brain scores should be accompanied by a systematic deconstruction of neural encoding performance, and an evaluation against simple and theoretically uninteresting features. Only after such deconstruction can we be somewhat confident that the neural predictivity of LLMs reflects core aspects of human linguistic processing. 

\section{Acknowledgements}
This work was supported by the following awards to JCK: NSF CAREER 1943467 and National Institutes of Health R01NS121097. (NSF CAREER supported EF; NIH R01 supported NH.)

\bibliography{references}

\begin{appendix}

\section{Appendix}

\subsection{Experimental data}
\label{sec:experimental_data}
\textbf{\textit{Pereira}:} For both experiments, each sentence was visually presented for $4$~s with $4$~s between sentences and an additional $4$ s between passages. A single fMRI scan was taken in the interval between each sentence. Because fMRI data is noisy, each experiment was repeated three times and fMRI data was averaged across the repetitions. A single fMRI scanning session consisted of $8$ runs, where each run contained $12$ passages in EXP1 and $9$ passages in EXP2. Participants performed a total of $3$ scanning sessions. The division of passages into runs and the order of the runs was randomized for each participant and scanning session.

\textbf{\textit{Fedorenko}:} Participants read sentence on word at a time, and each word was visually displayed for 450 or 700 ms. For each electrode, high gamma signal was extracted using gaussian filter banks at center frequencies ranging from $73-144$ Hz, the envelope of the high gamma signal was computed through a hilbert-transform, and the envelope was z-scored within each electrode. For each participant, language-responsive electrodes were selected where the z-scored envelope of the gamma activity was significantly higher during the sentences than a condition where participants read nonword lists. Z-scored high gamma activity from these language-selective electrodes were used in subsuquent analyses. 

\textbf{\textit{Blank}:} Text was split into $2$ s segments corresponding to each TR, with words that were on the boundary being assinged to the later TR. Due to the delay in the hemodynamic response function (HRF), neural activity was predicted using stimuli from $2$ TRs ($4$ s) previous. 

\textbf{Functional localization:} For \textit{Pereira} and \textit{Blank}, the language network was defined by the following procedure \citep{Fedorenko2011-kd}. First, voxels were identified in each participant which showed stronger responses to sentences compared to lists of non-words (sentences > non-word lists contrast). These voxels were then constrained by data-driven language activation maps formed by applying the same contrast to many other participants. Finally, the top $10\%$ of the voxels were selected which showed the greatest sentences > non-word lists difference. For \textit{Pereira}, we plot glass brain plots using model performance on four additional functional networks: multiple demand (MD), default mode network (DMN), auditory, and visual network. The multiple demand (MD) and default mode network (DMN) networks were defined using the same procedure, except that the contrast involved a spatial working memory task, where a hard > easy condition contrast was used for MD and a fixation > hard contrast was used for DMN \citep{Mineroff2018-xz}. Auditory and visual networks were defined using resting state connectivity \citep{Power2011-al}. 

\subsection{Banded ridge regression}
\label{sec:banded ridge}
We used a random search method to optimize the banded regression hyperparameters \citep{Dupre_la_Tour2022-sy}. Banded regression has two hyperparameters, $\gamma$, which is a vector of shape number of feature spaces that determines how much each feature space is scaled, and $\alpha$, which is the L2 penalty applied across feature spaces. Values for $\gamma$ are drawn from a Dirichlet distribution and hence sum to $1$. Down-scaling a certain feature space relative to others is functionally equivalent to assigning a separate L2 penalty for each feature space. This is because when a feature space is down-scaled, the L2 magnitude of the weights must increase for it to have a meaningful contribution to the predictions, which equates to increasing the L2 penalty for that feature space. The optimal $\gamma$ and $\alpha$ combination was found for each voxel/electrode/fROI by performing a random search over $\gamma$ values, storing the $\alpha$ value that performed best for that $\gamma$ on validation data, and then selecting the best performing $\gamma$ and $\alpha$ combination. We searched $40$ $\alpha$ values from $2^{-5}$ to $2^{34}$, exponentially spaced ($\alpha=0$ was also included). 

Before starting the random search, we tried all combinations of $\gamma$ values that removed feature spaces (i.e. down-scaled at least one feature space to $0$) to ensure the regression had an opportunity to remove features which hurt performance. In theory, this should obviate the need for the procedure implemented in \ref{sec:quantifying_overlap}. This is because the banded regression procedure can remove feature spaces based on validation data, meaning if a model performs worse than a sub-model the banded procedure has the opportunity to set the $\gamma$ value corresponding to the additional feature spaces to $0$. However, because neural data is noisy and there is often little data per subject, performance on validation data is not always indicative of performance on test-data. Therefore it is possible for the banded regression procedure to include a feature space (since it helps on validation data), and for this feature space to ultimately hurt test set performance, necessitating the correction procedure detailed in \ref{sec:quantifying_overlap}. 

We ran banded ridge regression for a maximum of $1000$ random search iterations with early stopping if the mean $R^2$ did not improve by more than $10^{-4}$  after $50$ iterations. We treated feature spaces with many dimensions as a single feature space because preliminary results showed this performed better. Specifically, we always treated the following feature spaces as one feature space: static word embeddings, sense-specific word embeddings, syntactic representations, and \textit{GPT2-XL} and \textit{RoBERTa-Large} best layer representations. Sentence length and sentence position were treated as one feature space, and word position and OASM were treated as separate feature spaces (i.e. given their own L2 penalty). We z-score all features across samples before training regressions, as is standard when using ridge regression in neural encoding studies.

\subsection{Additional details on train, validation, and test folds}
\label{sec:train_test}
\textbf{\textit{Pereira}:}  During each outer fold, a single passage from each of the $24$ semantic categories from one experiment was selected, and half of these passages were designated as the test set. This equated to $8$ test folds for experiment $1$ ($4$ passages per semantic category) and $6$ test folds for experiment $2$ ($3$ passages per semantic category). During each inner fold, we again selected one passage from each semantic category, and half of these passages were designated as validation (leading to $7$ inner folds for experiment $1$, and $5$  inner folds for experiment $2$). 

\textbf{\textit{Fedorenko}: }For each outer fold, we selected $4$ sentences as the test fold, resulting in $13$ outer folds. For each inner fold, we once again select $4$ sentences as the validation set, resulting in $12$ inner folds per outer fold. 

\textbf{\textit{Blank}:} For each outer fold, we selected a single story as the test fold, resulting in $8$ outer folds. For each inner fold, each of the remaining stories served in turn as the validation set, resulting in $7$ inner folds. 

\subsection{Orthogonal autocorrelated sequences model (\textit{OASM}) hyperparameters}
\label{sec:OASM}
The width of the Gaussian filter used for within-block smoothing was $\sigma=2.2$ in \textit{Pereira}, $\sigma=1.8$ in \textit{Fedorenko}, and $\sigma=1.5$ in \textit{Blank}. Gaussian widths were determined by sweeping $\sigma$ across $50$ evenly spaced values between $0.1$ and $5.0$ and choosing the best-performing $\sigma$ for each dataset.

\subsection{Shuffled train test splits confound task-relevant and task-irrelevant neural activity}
\label{sec:shuffled}
\textit{OASM} is a model which clearly lacks any linguistic representations that would allow it generalize to fully held-out passages. However, this is is not to say that \textit{OASM} is not correlated with linguistic features. For instance, sentences in a given passage are more semantically related with each other than with sentences in other passages.  Nonetheless, using shuffled train-test splits almost certainly exaggerates the variance explained by a model which, on the basis of semantic similarity, arrives at a similar representational structure as \textit{OASM}. This is because task-irrelevant neural responses make up a large fraction of neural activity \citep{Musall2019-sq}, and shuffled train-test splits allow a model with \textit{OASM}-like representational structure to predict not just the task-relevant neural responses driven by the participant reading the passage, but also any task-irrelevant neural activity that was present throughout the reading of the passage. Hence, we strongly urge researchers to avoid shuffled train test splits when evaluating the neural predictivity of language models, even when using decontextualized inputs as in \citep{Kauf2024-rh}, and we surmise that previous studies using shuffled train-test splits to compare neural predictivity between models might have come to erroneous conclusions.

\subsection{Across layer $R^2$ values in \textit{Pereira}}
\label{sec:across_layer}
\begin{figure}[h]
    \centering
    \includegraphics[scale=0.82]{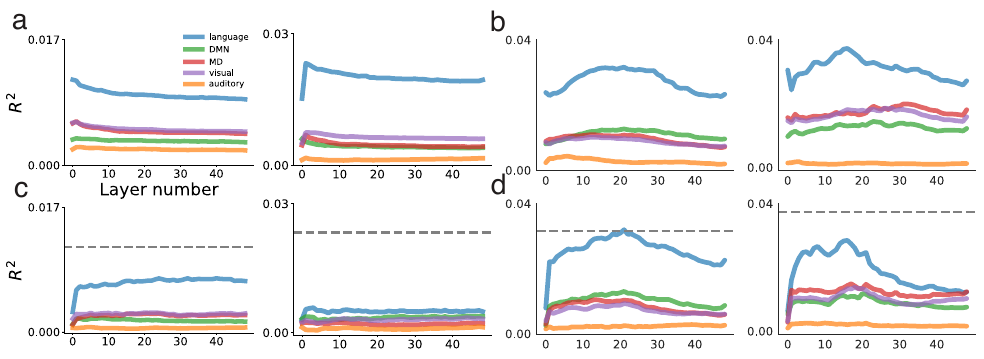}
 \caption{For all panels, EXP1 is on the left, and EXP2 is on the right. Layer $0$ is the input, static layer. (\textbf{a)}) Across layer performances in \textit{Pereira} for \textit{GPT2-XLU} for each functional network when using the sum-pooling method. \textbf{b)} Same as (\textbf{a}) but for \textit{GPT2-XL}, also using the sum-pooling method. \textbf{c)} Same as (\textbf{a}) but when using the last token method. Dotted grey line shows performance of best layer of \textit{GPT2-XLU} in language network when sum-pooling. \textbf{d)} Same as (\textbf{b}) but when using the last token method. Dotted grey line shows performance of best layer of \textit{GPT2-XL} in language network when sum-pooling.}
    \label{fig:across_layer_pereira}
\end{figure}

Across layer performances in \textit{Pereira} for \textit{GPT2-XLU} and \textit{GPT2-XL} when using the sum-pooling method (Fig. \ref{fig:across_layer_pereira}a,b) and the last token method (Fig. \ref{fig:across_layer_pereira}c,d). All $R^2$ values below $0$ were clipped to $0$ as done in previous analyses. Performance in language network is higher across the board than performance in DMN, MD, and visual networks. Furthermore, performance is lower with the last token method in every case except in EXP1 trained results where the last token method performs slightly better. Performance values were computed by averaging across all voxels in all participants within each functional network.

\subsection{Justification of statistical tests}
\label{sec:stats_test}
We define squared error values as $(y-\hat{y})^2$ , where $y$ is the neural data for a given voxel/electrode/fROI and $\hat{y}$ are the predictions from a model. To determine if one model, containing an LLM as a feature space, outperformed another model, which does not contain an LLM as a feature space, we performed a one-sided $t$-test between the squared error values from each model ($\alpha=0.05$).  While squared error values are not always normally distributed, our sample sizes were large (the minimum sample size was $243$) and so we still opted to use a $t$-test over a non-parametric alternative \citep{Lumley2002-lz}. 

We note that squared error values from a model are correlated, which means that the $t$-test is biased towards false positive results because correlated samples lead to an underestimation of the standard error of the mean. However, this does not impact the three main conclusions drawn from the \textit{t-}test in this study, namely: \textit{GTP2-XLU} does not account for significant additional variance over sentence length and sentence position in \textit{Pereira}, \textit{GTP2-XLU} does not significant explain significant additional neural variance over word position in \textit{Fedorenko}, and that \textit{GPT2-XL} predicts at chance levels on \textit{Blank}. This is because in all three cases, we do not find a voxel/electrode/fROI where squared error values from the model with an LLM as a feature space are significantly lower (i.e. a positive result) after FDR correction, and correlated samples only make a positive finding more likely to occur.

\subsection{Replication of trained \textit{GPT2-XL} results on \textit{Pereira} with \textit{RoBERTa-Large}}
\label{sec:roberta-large}

\begin{table}[h]
    \centering
    \caption{From left to right: model name, mean $R^2$ across participants after voxel correction, after voxel correction with \textit{RoBERTa-Large} (\textit{ROB}) (ensuring \textit{ROB} is included as a feature space), and the percentage of \textit{ROB} neural variance explained by the model ($\Omega\textsubscript{\textit{ROB}}(\textit{M}))$. Results are shown as EXP1, EXP2.}
    \begin{tabular}{cccc}
         \toprule
         \rule{0pt}{2ex} \textbf{Model} & \textbf{Voxel-corrected} & \textbf{w/ \textit{ROB}}& \textbf{$\Omega\textsubscript{\textit{ROB}}(\textit{M})$}\\ 
         \bottomrule 
         \hline 
         \rule{0pt}{2ex} SP+SL & $0.0147$, $0.0333$& $0.0347$, $0.0472$& $37.9 \pm 7.8$, $63.5 \pm 11.7$\\ 
         \hline
         +WORD & $0.0285$, $0.0478$& $0.0365$, $0.0535$& $81.4 \pm 4.9$, $86.9 \pm 5.0$\\ 
         \hline
         +SENSE & $0.0324$, $0.0508$ & $0.0379$, $0.05530$& $86.5 \pm 3.8$, $90.1 \pm 3.6$\\ 
         \hline
         +SYNT & $0.0354$, $0.0575$& $0.0396$, $0.0596$& $90.2 \pm 3.7$, $93.2 \pm 3.0$\\   
    \end{tabular}
    \label{tab:Trained results vc rob}
\end{table}

\begin{figure}[h]
    \centering
    \includegraphics[scale=0.95]{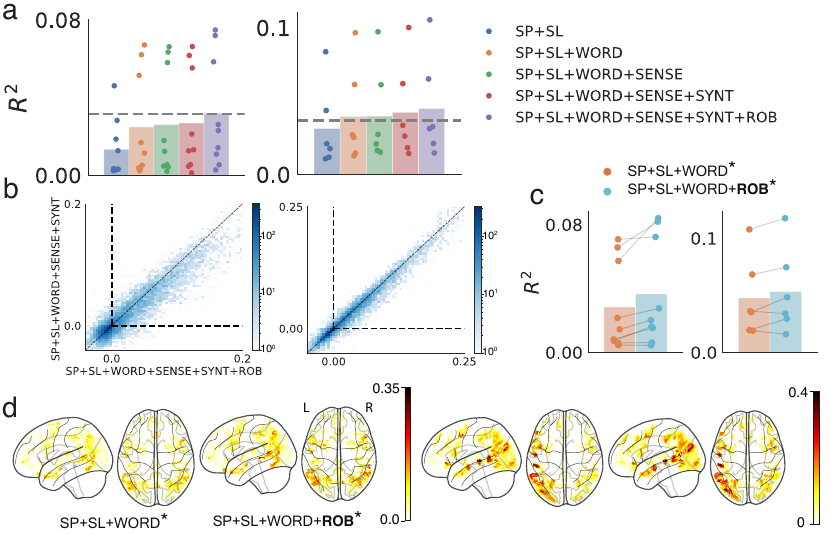}
    \caption{All panels are the same as Figure \ref{fig:trained_pereira}, except \textit{GPT2-XL }is replaced with \textit{RoBERTa-Large }(\textit{ROB}).}
    \label{fig:roberta-large}
\end{figure}

\begin{figure}[h]
    \centering
    \includegraphics[scale=0.8]{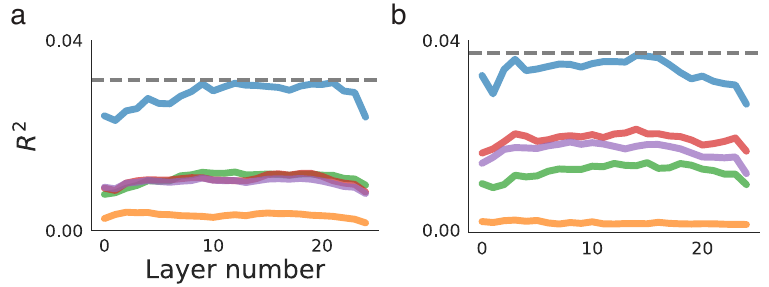}
    \caption{EXP1 is on the left, and EXP2 is on the right. Neural encoding performance, averaged across voxels, for each of the $24$ layers in \textit{RoBERTa-Large}. Colors denote functional networks, and color scheme is the same as Fig. \ref{fig:across_layer_pereira}. Dotted line indicates performance of the best layer of \textit{GPT2-XL} in language network}
    \label{fig:roberta-layer}
\end{figure}

To examine whether our results depend on the choice of LLM, we replicated all of our \textit{Pereira} trained analyses with \textit{RoBERTa-Large} (\textit{ROB}). The overall trend in results was the same as with \textit{GPT2-XL} (Fig. \ref{fig:roberta-large}). Namely, \textit{SP+SL+WORD} accounted for the majority of neural variance explained by \textit{ROB}, and \textit{SENSE} and \textit{SYNT} accounted for an additional, small amount (Table \ref{tab:Trained results vc rob}). The neural encoding performance of \textit{ROB} was highly similar to \textit{GPT2-XL}, with \textit{ROB} performing $98.3$\% and $98.8$\% as well as \textit{GPT2-XL} in the language network in EXP1 and EXP2, respectively (Fig. \ref{fig:roberta-layer}).

\subsection{Sense-specific word embeddings}
\label{sec:LMMS}
LMMS generates a sense embedding for each word by averaging across intermediate layer representations from an LLM for that sense of the word on a sense-annotated corpus. We used \textit{RoBERTa-Large} as our LLM of choice. Intermediate layers are weighted such that layers which are more helpful in word sense disambiguation are weighted more highly.  For word senses in WordNet where there are no annotated examples, LMMS sets their sense embedding equal to the average of other words in the same synset (i.e. synonyms). If there are no words in the same synset with sense embeddings, then LMMS tries to find words in the same hypernym. If that also fails, the sense embedding is set to the average of all sense embeddings from words in the same lexname. Finally, the sense embeddings are averaged together with the gloss embeddings for that sense of the word generated using \textit{RoBERTa-Large}. For additional details refer to \citet{Loureiro2022-ww}.

\subsection{Static word embeddings}
\label{sec:Static}
To maintain consistency with sense embeddings, we generate static word embeddings by taking a weighted average across all sense embeddings for a given word. The weight for each sense is proportional to its frequency count, as provided by WordNet version $3.0$ \citep{Miller1995-oj}. 

\subsection{Baseline model with 
\textit{GloVe} embeddings and no pronoun resolution}
\label{sec:GloVe}
In the \textit{Pereira} trained analyses, we defined the simpler, baseline model as sentence length, sentence position, and LMMS static word embeddings generated on pronoun dereferenced text. Since LMMS word embeddings are generated by averaging across intermediate layers of \textit{RoBERTa-Large}, we replicated our baseline model results using static word embeddings from \textit{GloVe}, which is a simpler method that is not dependent on the internal representations of a transformer. Analagous to the LMMS static word embeddings, we first generated \textit{GloVe} sentence embeddings by obtaining \textit{GloVe} vectors on content words on pronoun-dereferenced text, and sum-pooling vectors within each sentence. To evaluate the contribution of pronoun resolution in explaining \textit{GPT2-XL} neural encoding performance, \textit{GloVe} embeddings were generated using the same procedure on text that was not pronoun-dereferenced. 
\begin{figure}[h]
    \centering
    \includegraphics{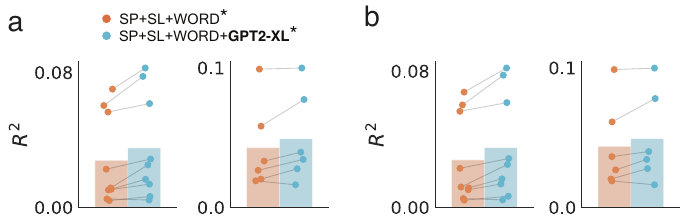}
    \caption{EXP1 is on the right, and EXP2 is on the left. The \textit{WORD} feature space is composed of \textit{GloVe} embeddings generated on pronoun-dereferenced text in \textbf{(a)} and on text that is not pronoun-dereferenced in \textbf{(b)}.}
    \label{fig:glove}
\end{figure}

\subsection{Contextual syntactic representations}
\label{sec:SYNT}
Syntactic embeddings are derived by substituting content words (nouns, proper nouns, verbs, adjectives, and adverbs) in the original sentences with words of matching part-of-speech and dependency tag via the SpaCy transformer-based model \citep{spacy2}. First, we create a word bank for each part-of-speech and dependency tag by passing $300,000$ sentences from the Generics KB corpus \citep{huggingface:dataset} through the SpaCy transformer-based model. Then, we run each sentence in \textit{Pereira} through the SpaCy transformer-based model. We then generate new sentences, in which each content word from the original sentences is replaced with a content word of matching part-of-speech and dependency tag randomly sampled from the word bank. Each newly generated sentence is then once again passed through the SpaCy transformer-based model to get the token indices of the subtrees associated with each token; we ensure that, for each token index, the token indices of the subtree in the generated sentence matches that in the original sentence. If not, we discard the generated sentence and try again. We continue until $170$ new sentences with valid subtrees are generated for each original sentence. Finally, we take the cross-entropy loss of \textit{GPT2-XL} on each newly generated sentence and keep the $100$ sentences with the smallest loss.

To get a syntactic representation for each original sentence, we run these $100$ new sentences through \textit{GPT2-XL}, take the sum-pooled representation within each of them, and then take the mean representation across them. We take the syntactic representation from the same best-performing layer of \textit{GPT2-XL} that we ultimately compare it to.

This method is highly similar to that of \citep{Caucheteux2021-nv}, the main difference being that we keep function words from the original sentence. We do this for two reasons: (i) representations of function words have previously been used to generate representations of syntactic form for neural encoding \citep{Kauf2024-rh} and (ii) coherent sentences were generated much more often when function words were left intact. Altogether, our syntactic representations combine the constraints on part-of-speech and dependency structure, as in \citep{Caucheteux2021-nv}, and the preservation only of function words, as in \citep{Kauf2024-rh}.

\subsection{Word position}
\label{sec:FedWP}

\begin{figure}[h]
    \centering
    \includegraphics[scale=0.3]{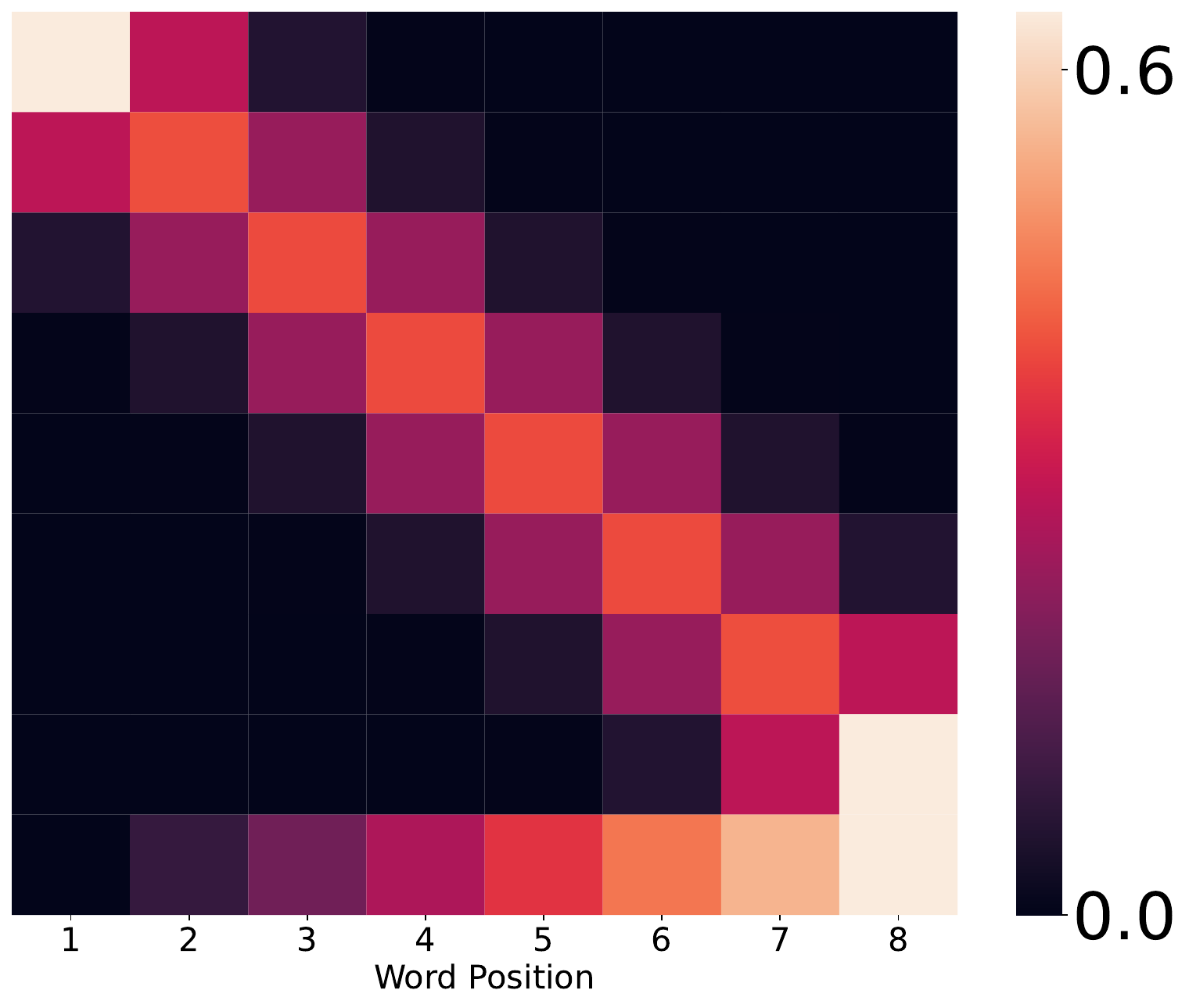}
    \caption{Word Position feature for a single sentence in \textit{Fedorenko}.}
    \label{fig:fed_wp}
\end{figure}
The primary finding in the paper which first collected \textit{Fedorenko} \citep{Fedorenko2016-eg} was a ramping of neural activity across the words of sentences, where each sentence was $8$ words long.  Hence, we concatenate a linearly ramping $1$-dimensional positional signal to an $8$-dimensional $1$-hot positional signal. Because we expect positional representations in the brain to be more similar between adjacent words than more distant words, we apply a Gaussian filter ($\sigma=1$) to the $8$-dimensional positional signal.  The resulting feature space, which we refer to as "word position" in the main text, is shown for a single sentence in Figure \ref{fig:fed_wp}

\subsection{\textit{OASM} and \textit{GPT2-XL} Model Comparison on \textit{Blank}}
\label{sec:BlankOASM}

\begin{figure}[h]
    \centering
    \includegraphics[scale=0.3]{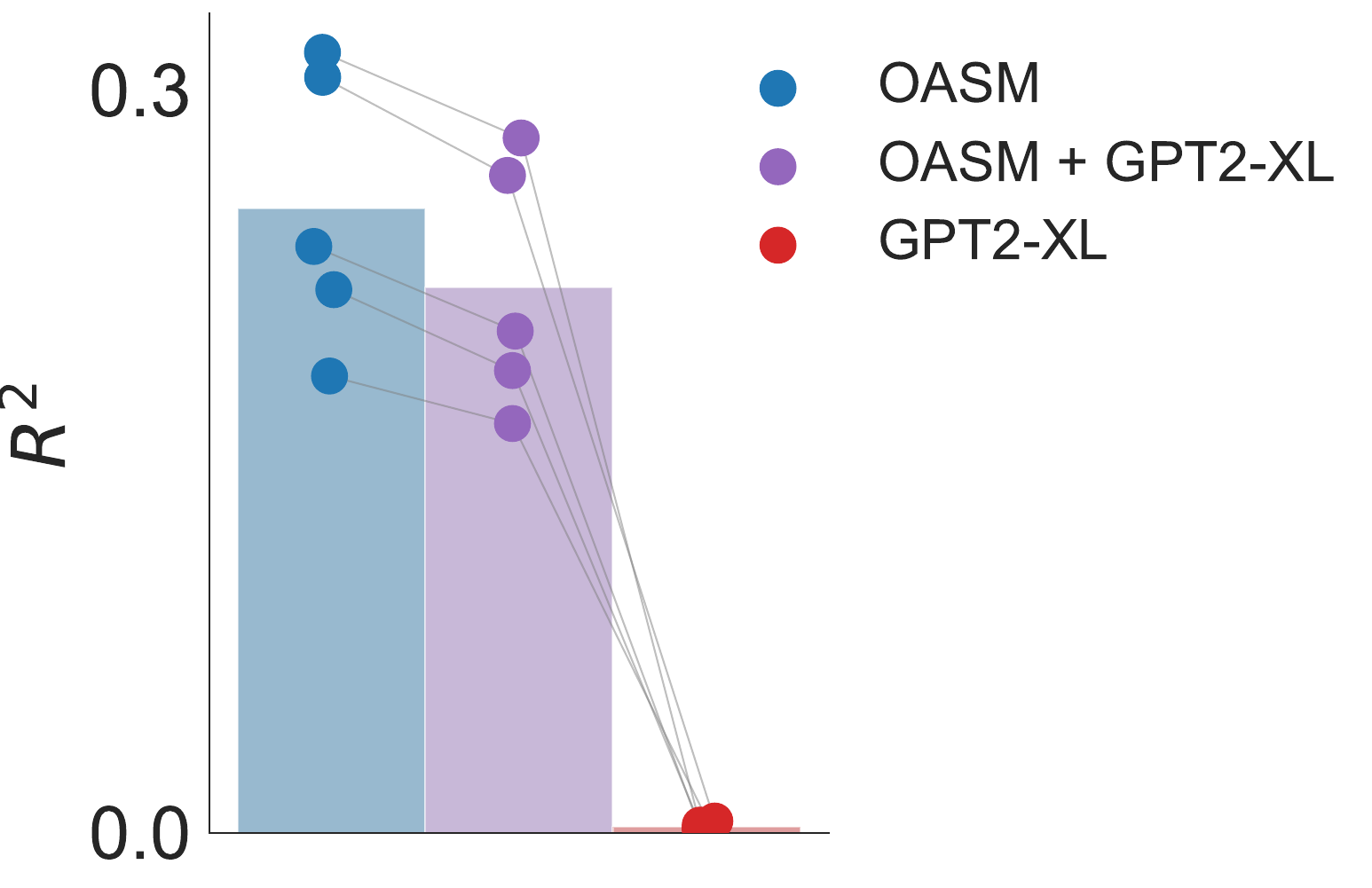}
    \caption{\textit{OASM} far outperforms \textit{GPT2-XL} on \textit{Blank}, and \textit{GPT2-XL} does not appear to explain any variance beyond that explained by \textit{OASM}.}
    \label{fig:blank_res}
\end{figure}
We find that \textit{OASM} achieves $103.6$ times higher neural predictivity than \textit{GPT2-XL} on \textit{Blank} when using shuffled train-test splits. There could be several reasons for this. First, it might be that the method for pooling representations from \textit{GPT2-XL} used here (\ref{llm feature pooling}) and in \citep{Schrimpf2021-pg, Hosseini2024-kg, aw2024instructiontuned} did not yield useful enough representations for \textit{GPT2-XL} to map effectively to the brain data. An additional likely culprit is that, of the three datasets we study here, \textit{Blank} has the greatest potential for autocorrelation in temporally adjacent samples. This is because, while \textit{Pereira} typically has a TR every $8$ seconds, \textit{Blank} has a TR every $2$ seconds. Lastly, our results here are not completely surprising; given that \citep{Schrimpf2021-pg, Hosseini2024-kg} observed \textit{GPT2-XLU} models perform better than \textit{GPT2-XL} on this dataset, it did not seem likely that \textit{GPT2-XL} would map onto neural representations of linguistic features here.

\subsection{Computational Resources}
All analyses were done between $2$ machines. The first machine has $1$ RTX 4090 GPU, and the second machine has $1$ RTX 3090 GPU.  The most computationally demanding parts of our analyses were fitting the banded ridge regressions used to generate Figure \ref{fig:trained_pereira}, collecting untrained model results across $10$ seeds, and generating syntactic representations, which each took around $3$ hours to complete.

\subsection{Dataset Licenses}
\textit{Blank} was originally released as part of the Natural Stories Corpus, which is provided under the CC BY-NC-SA license \citep{Futrell2018-sk}. \textit{Pereira} is released under the Creative Commons License \citep{Pereira2018-ry}. The version of \textit{Fedorenko} used here is provided under the MIT license. All datasets used are the same versions as in \citep{Schrimpf2021-pg} and can be downloaded using the \href{https://github.com/mschrimpf/neural-nlp/tree/master}{neural-nlp} repository. All datasets were collected with IRB approval at their respective institutions. The Generics KB corpus used to generated syntactic representations is available under the Creative Commons license. 
\end{appendix}

\subsection{Code}
All code used in this paper can be found at \href{https://github.com/ebrahimfeghhi/beyond-brainscore}{beyond-brainscore} on GitHub.

\end{document}